%% file: root.tex
\documentclass[letterpaper, 10 pt, conference]{ieeeconf} %
\usepackage{algorithm}
\usepackage{algorithmic}
\usepackage{pgfplots}
\usepackage{capt-of}
\usepackage{caption}
\usepackage{subcaption}
\usepackage{cite}
\usepackage{bm}
\usepackage{amsfonts}
\usepackage{hyperref}
\let\labelindent\relax
\usepackage{enumitem}
\usepackage{multirow}
\usepgfplotslibrary{groupplots}
\pgfplotsset{compat=1.18}

\makeatletter
\newcommand\fs@spacedruled{%
    \def\@fs@cfont{\bfseries}%
    \let\@fs@capt\floatc@ruled
    \def\@fs@pre{\kern3.5mm\hrule height.8pt depth0pt \kern2pt}%
    \def\@fs@mid{\kern2pt\hrule\kern2pt}%
    \def\@fs@post{\kern2pt\hrule\relax}%
    \let\@fs@iftopcapt\iftrue
}
\floatstyle{spacedruled}
\restylefloat{algorithm}
\makeatother

\definecolor{epcol}{HTML}{4C78A8}
\definecolor{eucol}{HTML}{F58518}
\definecolor{valcol}{HTML}{54A24B}
\definecolor{ourscol}{HTML}{B279A2}

\IEEEoverridecommandlockouts %

\usepackage{amsmath} %
\usepackage{xcolor}
\usepackage{comment}
\usepackage{booktabs}
\usepackage{todonotes}
\usepackage{pgfplots}
\pgfplotsset{compat=1.18}
\usepackage{cleveref}

\usepackage{xspace}
\newcommand{\method}{FCTS\xspace}

\makeatletter
\renewcommand\subsubsection{\@startsection{subsubsection}{3}{\parindent}%
{0ex plus 0.1ex minus 0.1ex}{0ex}{\normalfont\normalsize\bfseries}}
\makeatother

\usepackage{eso-pic}
\AddToShipoutPictureFG*{\AtPageUpperLeft{\raisebox{-0.42in}[0pt][0pt]{\makebox[\paperwidth]{\footnotesize\begin{minipage}{0.82\paperwidth}\centering
This work has been submitted to the IEEE for possible publication.\\
Copyright may be transferred without notice, after which this version may no longer be accessible.
\end{minipage}}}}}

\title{\LARGE \bf
\method: Failure-Guided Curriculum Tree Search\\Over Generated Terrain for Legged Parkour
}

\author{Narayanan Palghat Parameswaran$^{1,2^*}$,
Dong-Ki Kim$^{2}$,
Jay Patrikar$^{2}$,
Santiago Gutierrez$^{2}$, \\
Harshad Zade$^{2}$,
Shayegan Omidshafiei$^{2}$,
Harmish Khambhaita$^{2}$
\\
\\ $^{1}$Carnegie Mellon University $^{2}$FieldAI
\\ \thanks{$^{*}$ This work was completed when Narayanan was an intern at FieldAI}}

\begin{document}

\maketitle
\thispagestyle{empty}
\pagestyle{empty}

\input{sections/abstract}

\input{sections/introduction}
\input{sections/related_works}
\input{sections/method}
\input{sections/experiments}

\input{sections/conclusions}

\bibliographystyle{IEEEtran}
\bibliography{bibtex.bib}

\end{document}

%% file: sections/abstract.tex
\begin{abstract}

Quadruped parkour policies handle canonical obstacles well, but their training curricula often miss the irregular, compound geometry of industrial and disaster-response sites, where agile locomotion matters most. Such environments pose a hard generalization problem: even training directly on reconstructed geometry from the target site does not reliably yield policies that traverse it.
Hand-designed curricula span only a few terrain families, and recent LLM-generated curricula offer little control over which policy weaknesses they target.
We propose \method, a failure-conditioned terrain curriculum framework that uses localized traversal failures to automatically generate the terrains a policy should train on next. Each failure's location, termination mode, robot state, and surrounding terrain geometry condition this generation, and Monte Carlo Tree Search explores alternative curriculum trajectories under a fixed training budget.
On terrains built from scanned real-world construction sites, \method raises success rate from 28.4\% to 85.8\% on a medium-difficulty course and from 14.9\% to 44.7\% on a hard course relative to the strongest parkour or LLM-curriculum baselines, and maintains higher progress on extreme terrains where full traversal is rare.
Hardware experiments on a quadruped robot confirm that the learned \method policies transfer to the real world. Please see our \href{https://narayananpp.github.io/fcts/}{Project Page} for additional videos.

\end{abstract}

%% file: sections/introduction.tex
\section{INTRODUCTION}

\begin{figure}[t]
    \centering
    \includegraphics[width=0.95\columnwidth]{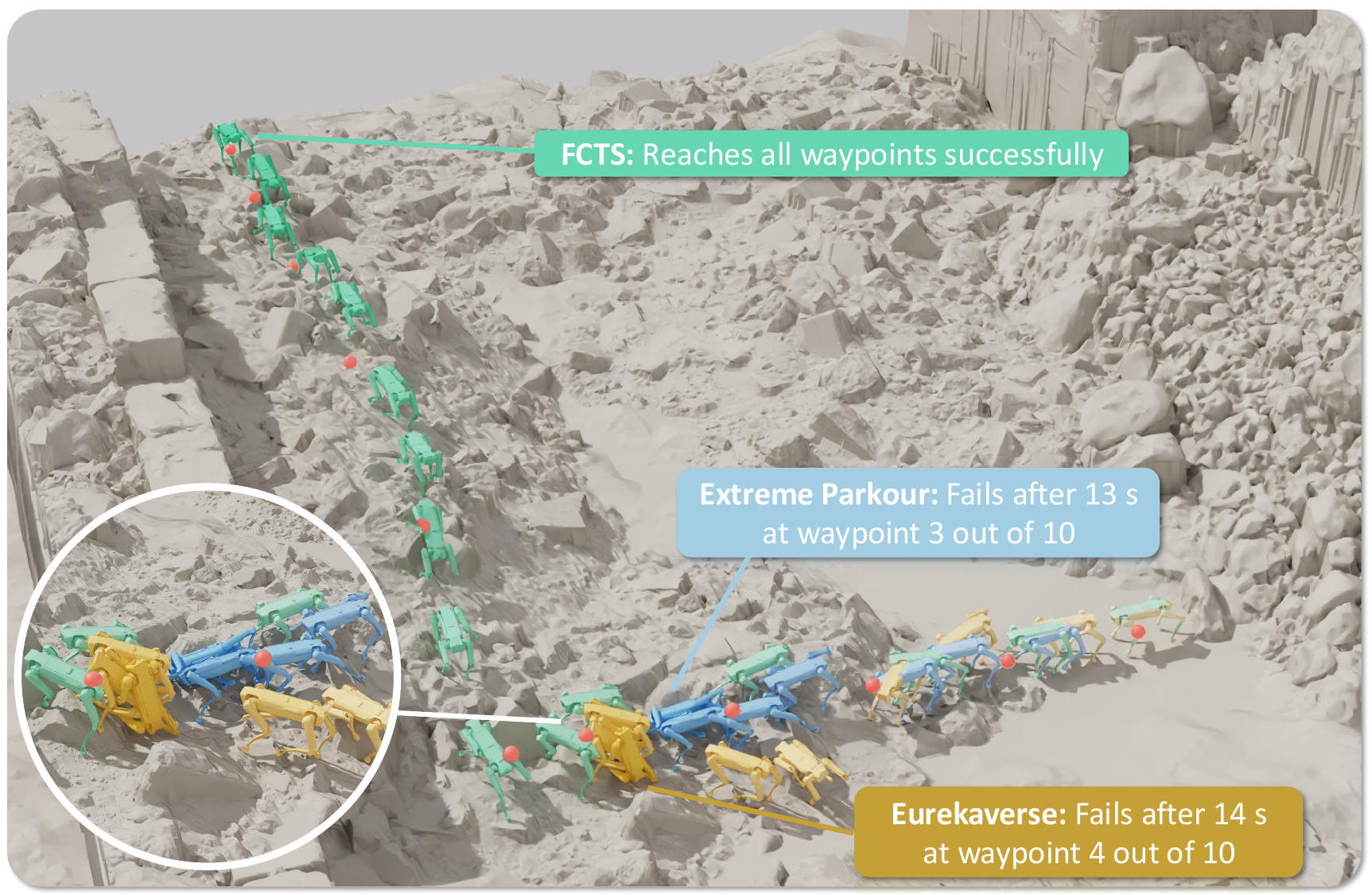}
    \caption{\textbf{\method traverses reconstructed real-world terrain on which
    baseline curriculum stop early.}
    The site is an unseen deployment environment, held out from curriculum
    generation and training, and all three policies start from the same
    pre-trained checkpoint under the same commanded route, shown as overlaid
    poses through waypoints marked in red.
    Each trajectory is a representative episode.
    \label{fig:teaser}}
\end{figure}

Learning-based quadruped locomotion has enabled robots to traverse gaps, stairs, hurdles, and elevated platforms with remarkable agility~\cite{dreamwaq,robotparkour,anymalparkour,extremeparkour,soloparkour}.
However, these capabilities are typically learned from a pre-defined training distribution of terrain with carefully controlled obstacle variations.
Such policies often break down on real-world environments like those in \Cref{fig:teaser}.
Irregular surfaces, discontinuities, and narrow supports are each familiar in isolation, but they co-occur in arrangements that fall outside this distribution.
The training distribution, not the policy, is the binding constraint.

Traditional legged-robot training relies on manually designed terrain families
that define the obstacle structures encountered during learning.
Methods such as Extreme Parkour (EP)~\cite{extremeparkour},
Robot Parkour Learning~\cite{robotparkour}, and
ANYmal Parkour~\cite{anymalparkour} construct such terrain families and use a
difficulty scheduler to increase terrain difficulty as the policy improves.
However, the underlying terrain families themselves remain fixed throughout
training. Extending this terrain library by hand is difficult, time-consuming, and can be sub-optimal. This motivates generating new training terrains
automatically from the policy's observed failures.

Recent work has begun to use Large Language Models (LLMs) to automate the design of
executable components for reinforcement learning. Eureka~\cite{eureka}
demonstrated that an LLM can generate and iteratively improve reward code, often
outperforming expert-designed rewards across a broad set of tasks.
Building on this idea, Eurekaverse~\cite{eurekaverse} applies LLM-based
generation directly to legged-robot training terrains and synthesizes executable
terrain programs and revises them based on policy performance.

\textbf{What signal drives generation?}
Eurekaverse evaluates the policies trained on its generated
terrains and uses the best-performing policy and its associated terrain set to
guide the next round of generation. The feedback is therefore an aggregate
performance signal: it indicates how the policy performed \textbf{on average} on the generated terrain set,
but not where traversal failed or which local geometry caused the failure.
Without this localized information, subsequent terrain generation follows the LLM’s prior over
plausible terrain rather than the policy’s weaknesses. What is missing is \textit{feedback that identifies where the
policy fails and why}.

Our key insight is that a failed traversal provides a rich diagnostic signal: it \textit{localizes where the policy's competence breaks down and characterizes the conditions at that point} using cues such as failure coordinates, robot state at termination, local terrain geometry, and visual context.
These specify what terrain to train on next, so that each expansion measures that boundary and moves it toward the target domain.

We formulate curriculum discovery in a principled way by building on the
Monte Carlo Tree Search framework~\cite{coulom2006mcts}: checkpoints form nodes,
while edges correspond to RL training runs on LLM-generated terrains conditioned
on failure feedback of the parent checkpoint.
Each node in the search tree rehearses terrain from its own ancestor nodes in the branch to limit forgetting.
Under a fixed wall-clock budget, the search returns a portfolio of policy
checkpoints.

We refer to this method as \textbf{\method}, for \emph{Failure-Guided Curriculum Tree Search}.
Our contributions are:
\begin{itemize}[leftmargin=*]
    \item \textbf{Failure-Conditioned Terrain Curriculum Generation.}
    We close the generation loop on observed failures rather than on an aggregate
    score, conditioning each new terrain on the failure location, termination
    condition, robot state, and surrounding geometry.

    \item \textbf{An MCTS Formulation of Automated Terrain Curriculum Generation.}
    We cast curriculum discovery as a tree search over policy checkpoints and
    terrain-conditioned training runs, with branch-history rehearsal and anytime
    search under a fixed wall-clock budget.

    \item \textbf{Evaluation on Reconstructed Real-World Sites.}
    We evaluate on held-out courses constructed from scanned industrial and
    disaster-response geometry, and demonstrate transfer to a physical quadruped.
\end{itemize}

%% file: sections/related_works.tex
\section{RELATED WORK}

\textbf{Quadruped Parkour Learning.}
Learning-based locomotion has enabled quadruped robots to traverse increasingly challenging and diverse terrain. Robot Parkour Learning~\cite{robotparkour} learns parkour skills including climbing, leaping, crawling, and running, and distills them into a unified parkour policy. ANYmal Parkour~\cite{anymalparkour} learns specialized locomotion skills and selects among them to navigate complex obstacle courses. More recent approaches explore end-to-end visual control~\cite{extremeparkour} and constrained reinforcement learning from privileged experience~\cite{soloparkour}. These methods demonstrate capable parkour policies, but their training environments are constructed from terrain families and obstacle distributions specified in advance. Our work instead focuses on automatically discovering training curricula as the capabilities of the policy evolve.

\textbf{Automatic Curriculum and Environment Design.}
Automatic curriculum learning adapts the training distribution according to the current capabilities of an agent. GoalGAN~\cite{goalgan} generates goals near the frontier of the agent's ability, while ALP-GMM~\cite{alpgmm} samples environment parameters according to learning progress. Prioritized Level Replay~\cite{plr} prioritizes previously encountered levels according to their estimated learning potential. Unsupervised Environment Design extends this idea by actively constructing training environments. PAIRED~\cite{paired} generates environments through a regret-based adversarial objective, while ACCEL~\cite{accel} gradually evolves complex environments near the frontier of agent capability. POET~\cite{poet} jointly evolves populations of environments and their associated solutions, and explores multiple environment--policy trajectories. These methods establish the effectiveness of adapting the training environment to the learner, but typically operate over parameterized or procedurally defined environment spaces.
Closest to our setting, TACL~\cite{tacl} learns a policy-conditioned difficulty
function over pre-existing terrain maps. In contrast, we generate new terrain
geometry from localized policy failures.

\begin{figure*}[!t]
    \centering
    \vspace{2mm}
    \includegraphics[width=0.98\textwidth]{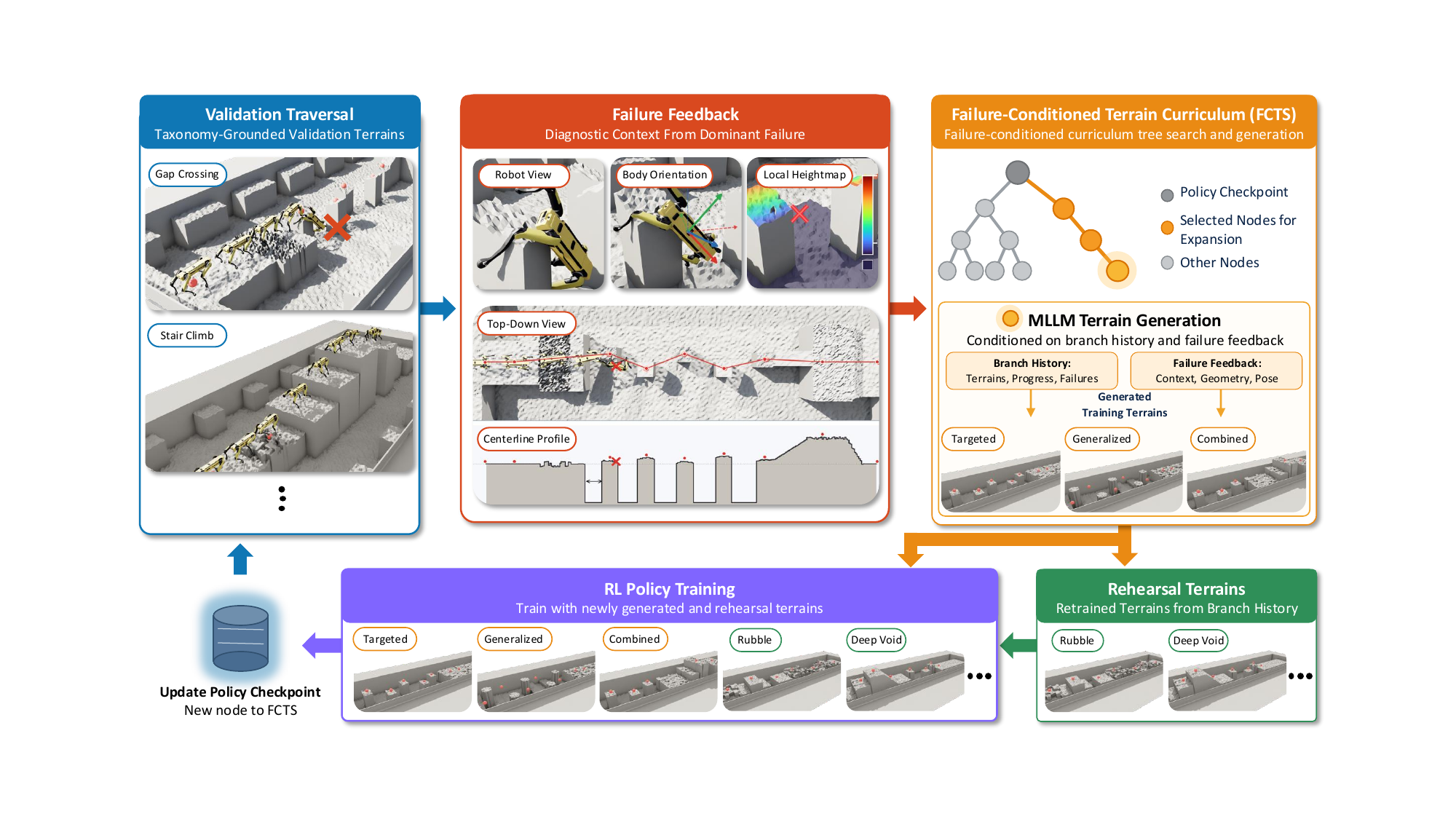}
    \caption{
    Overview of our failure-conditioned curriculum search (FCTS). Validation failures
    guide LLM terrain generation, while parallel MCTS explores alternative
    policy-improvement trajectories.
    }
    \label{fig:framework}
    \vspace{-1.5mm}
\end{figure*}

\textbf{LLM-Guided Curriculum Generation and Search.}
LLMs provide a more expressive mechanism for automatic curriculum design through their reasoning and code-generation capabilities.
EnvGen~\cite{envgen} generates training environments with an LLM and adapts them using agent-performance feedback to improve weak skills.
Most closely related to our setting, Eurekaverse~\cite{eurekaverse} uses LLMs to generate and evolve executable terrain programs for quadruped parkour. These works demonstrate that LLMs can naturally expand the space of environments considered during training. Our work complements this capability with detailed, localized feedback from robot traversal failures and a structured search over successive curriculum improvements. Tree search has also been combined with LLM code generation: GIF-MCTS~\cite{gifmcts} uses MCTS to generate, improve, and repair code world models, and MCTS has been applied to search over RL update sequences~\cite{mctsrl}. These searches score each candidate by cheaply executing the generated program. In our setting each node is a policy checkpoint and each edge is an RL training run, so tree search operates over policy improvement trajectories rather than over programs.

%% file: sections/method.tex
\section{DISCOVERING TERRAIN CURRICULA\\FROM FAILURES}

We now present \method in full, first fixing the setup and the basis on which policy checkpoints are measured, then describing the search loop that uses it. \Cref{fig:framework} summarizes the search loop and \Cref{alg:fcts} states it as pseudocode.

\textbf{Overview.}
\method is an outer-loop search wrapped around an unmodified parkour learner: the learner turns a set of terrains into a policy, and \method decides which terrains it receives next and which checkpoint receives them.
Deciding that requires three things the inner learner cannot supply.
It needs a fixed basis on which two checkpoints trained on different terrain can be compared at all, a description of a checkpoint's failures detailed enough to specify geometry rather than merely score it, and a rule for spending an expansion when each one costs a complete training run and cannot be undone.

\subsection{Preliminaries}

\subsubsection{Problem setup}
\label{subsec:problem}

Let $\pi_0$ denote an initial pre-trained parkour policy and $J(\pi)$ the validation performance of a policy under a common evaluation criterion.
Our goal is to improve $\pi_0$ through a sequence of terrain curricula such that the resulting policy can traverse complex real-world environments.
A training run applies the underlying reinforcement-learning parkour procedure $\mathcal{T}$ to a checkpoint and a curriculum:
\begin{equation}
    \pi_j=\mathcal{T}(\pi_i,\mathcal{C}_{i}),
    \label{eq:training_run}
\end{equation}
where $\mathcal{C}_{i}$ is the curriculum generated for the current expansion of $s_i$; repeated expansions of $s_i$ generate different curricula and therefore different children.
Curricula are not interchangeable, because the effectiveness of one depends on the capabilities of the policy it starts from: the same terrain that advances a weak checkpoint may teach a strong one nothing.

What makes this a search problem rather than a design problem is cost.
Each application of $\mathcal{T}$ is a complete training run, so only a handful of curricula can ever be tried.
Given a budget of $N_{\mathrm{train}}$ runs, our objective is:
\begin{equation}
    \pi^{*}
    =
    \arg\max_{\pi \in \mathcal{P}_{N_{\mathrm{train}}}}
    J(\pi),
    \label{eq:objective}
\end{equation}
where $\mathcal{P}_{N_{\mathrm{train}}}=\{\pi_0,\pi_1,\dots,\pi_{N_{\mathrm{train}}}\}$ is the set of checkpoints discovered within that budget, the root policy together with every checkpoint produced by a completed run.
Because the budget binds, the question is not which curriculum is good in isolation but which checkpoint deserves the next run and what terrain that run should use.
\method answers it by scoring candidates on a fixed validation basis, conditioning new terrain on where each candidate failed, and spending what remains of the budget on the branches that have improved the most.

\subsubsection{Validation terrain design}
\label{subsec:validation}

Checkpoints in the tree are trained on different terrain, so nothing about their training distributions makes their performance comparable.
The search therefore needs one fixed yardstick, held constant for the whole run, or branch values measured at different depths mean different things and selection becomes meaningless.
We use an LLM conditioned on standard parkour terrain taxonomies~\cite{extremeparkour, robotparkour, anymalparkour, zhang2026ame} to generate this set, covering challenges such as box-climbing, gap-leap, narrow-support walking, and off-camber traversal, so that the yardstick spans the axes a parkour policy can fail along rather than a single difficulty scale.
The same evaluation serves twice: it produces the common validation signal that makes branches comparable, and it produces the localized failure feedback that tells the generator what to build.

\subsection{Failure-Guided Curriculum Tree Search}

An expansion is expensive and irreversible, and it changes the failures the search will observe afterwards, so the space cannot be resampled and a greedy sequence commits to one trajectory through it.
\method therefore keeps every expansion it has paid for in a persistent tree and treats the choice of where to expand next as the decision worth making carefully.
Each node $s_i$ holds a retained checkpoint $\pi_i$ and a visit count $N(s_i)$; $Q(s_i,s_j)$ holds the value of the branch from parent $s_i$ to child $s_j$; and each node admits at most $B$ children, so a path through the tree is an ordered sequence of checkpoints together with the curricula that produced them.
One iteration follows the standard Monte Carlo tree search phases, with one addition: \method inserts a \emph{failure diagnosis} phase between selection and expansion, and it is that phase which lets terrain be conditioned on a specific failure rather than on a score.
The phases below track \Cref{alg:fcts}.

\begin{algorithm}[!t]
\caption{\method: Failure-Conditioned Curriculum Search}
\label{alg:fcts}

\small
\begin{algorithmic}[1]

\REQUIRE Initial policy $\pi_0$, validation terrains $\mathcal{V}$,
training budget $T_{\max}$

\STATE Initialize search tree with root $\pi_0$

\WHILE{training budget remains}

    \STATE Select checkpoint $\pi_i$ using PUCT
    \STATE Evaluate $\pi_i$ on $\mathcal{V}$
    \STATE Collect failure feedback $\mathcal{F}_i$ and branch history $\mathcal{L}_i$
    \STATE Select rehearsal terrain set $\mathcal{R}_i$ from $\mathcal{L}_i$

    \STATE Generate curriculum $\mathcal{C}_{i}$ conditioned on
    $(\mathcal{F}_i,\mathcal{L}_i)$

    \STATE Train $\pi_i$ on $\mathcal{C}_{i}$ and rehearsal terrains
    $\mathcal{R}_i$ to obtain $\pi_j$

    \STATE Add $\pi_j$ as a child of $\pi_i$
    \STATE Estimate $V(\pi_j)$ using Algorithm~\ref{alg:rollout}
    \STATE Back up $V(\pi_j)$ along the selected tree path

\ENDWHILE

\RETURN search tree

\end{algorithmic}
\end{algorithm}

\subsubsection{Selection (line~3)}
The search must neither abandon a branch that is improving nor keep exploiting it while untried siblings exist, so we select with the PUCT rule~\cite{Silver2016MasteringTG}:
\begin{equation}
    \mathrm{PUCT}(s_i,s_j)
    =
    Q(s_i,s_j) + U(s_i,s_j),
\end{equation}
whose second term rewards branches that have been visited rarely relative to their parent:
\begin{equation}
    U(s_i,s_j)
    =
    c_{\mathrm{puct}} P(s_i,s_j)
    \frac{\sqrt{N(s_i)}}{1+N(s_j)} .
\end{equation}
Here, $P(s_i,s_j)$ is the prior on child $s_j$, $N(s_i)$ and $N(s_j)$ are the parent and child visit counts, and the scalar $c_{\mathrm{puct}}\in\mathbb{R}$ sets how strongly exploration is weighted.
A learned prior is unavailable here, because the children of a node are terrain curricula that do not exist until the LLM writes them and so cannot be scored in advance.
We therefore take the prior uniform and absorb the resulting constant into $c_{\mathrm{puct}}$, which leaves:
\begin{equation}
    S(s_i,s_j)
    =
    \underbrace{Q(s_i,s_j)}_{\mathrm{exploitation}}
    +
    \underbrace{
    c_{\mathrm{puct}}
    \frac{\sqrt{N(s_i)}}{1+N(s_j)}
    }_{\mathrm{exploration}},
    \label{eq:puct}
\end{equation}
and at each node the search descends into:
\begin{equation}
    s_j^{*}
    =
    \arg\max_{s_j \in \mathrm{Child}(s_i)}
    S(s_i,s_j),
\end{equation}
applied recursively from the root until it reaches the node $s_i$ to expand.

\begin{figure*}[!t]
    \centering
    \vspace{2.5mm}

    \hspace*{0.025\textwidth}
    \includegraphics[width=0.45\textwidth]{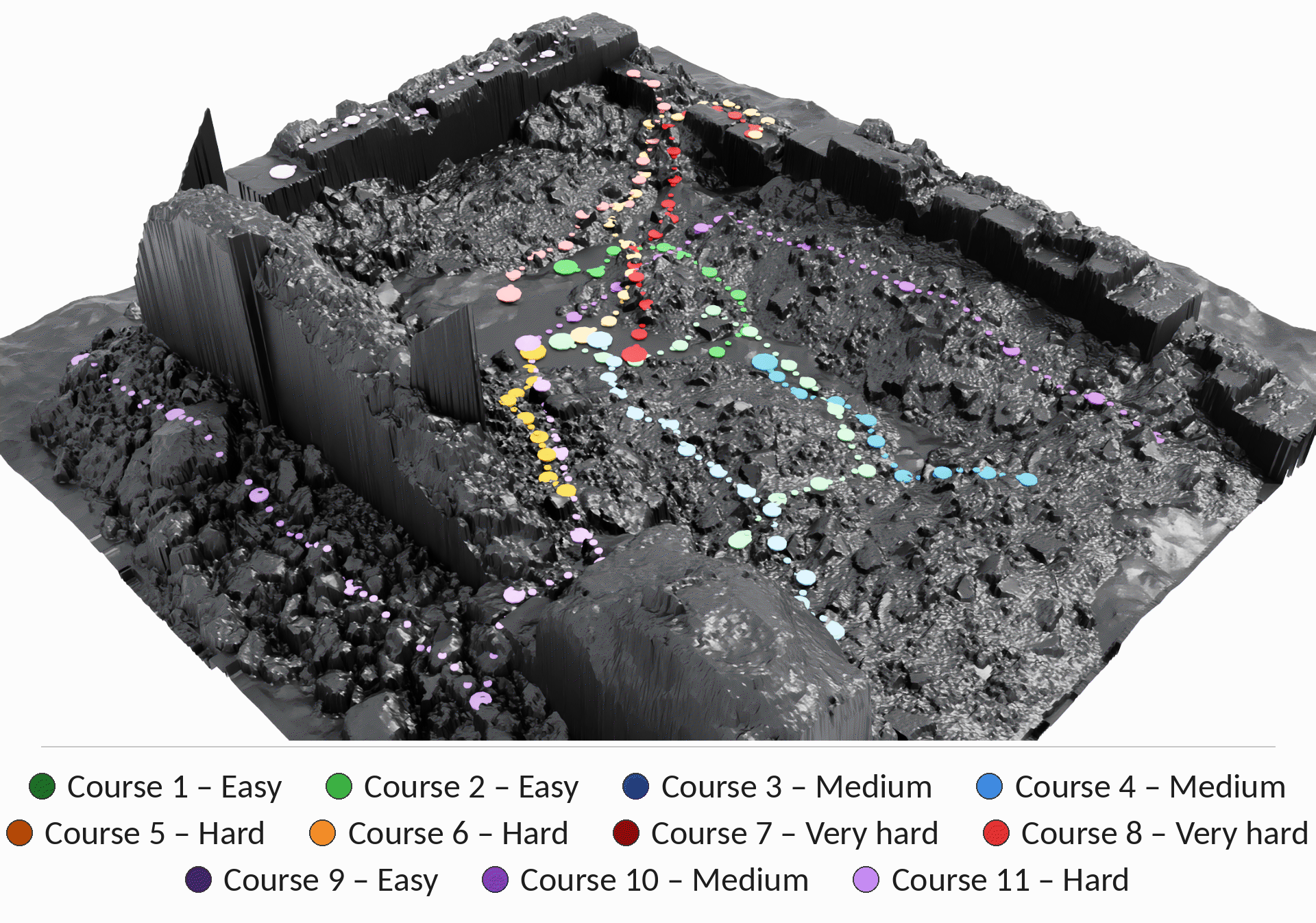}
    \hfill
    \includegraphics[width=0.45\textwidth]{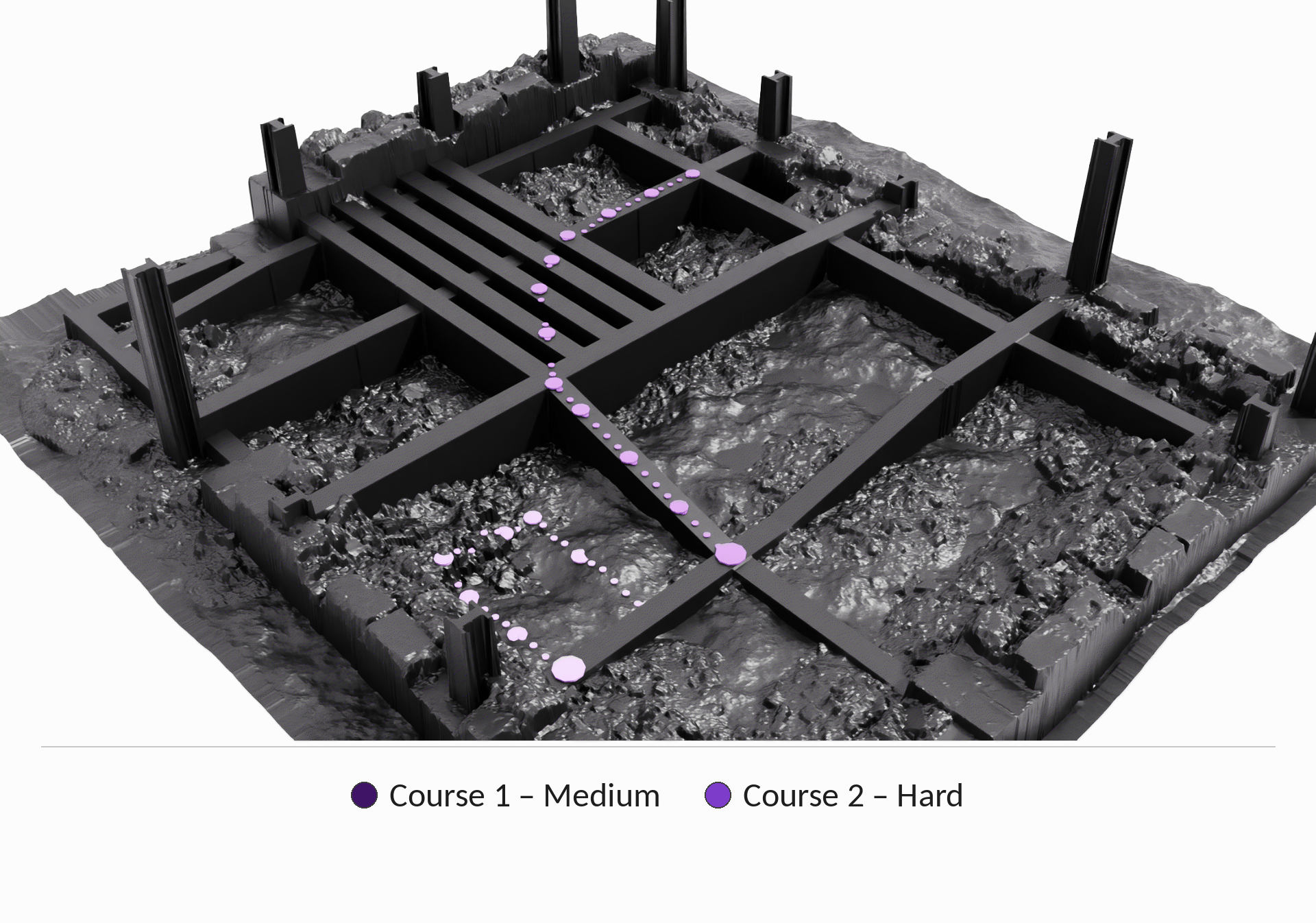}
    \hspace*{0.025\textwidth}

    \caption{
    \textbf{Evaluation routes in the two reconstructed environments.}
    \textbf{Left (Debris site)} shows both the relief-stratified traversal routes
    (Courses 1--8) and directly specified traversal routes (Courses 9--11)
    across all difficulty levels.
    \textbf{Right (Exposed Beam site)} shows the directly specified evaluation
    routes only, since the relief-based course generator is not reliable on its
    large gaps and disconnected structures.
    }
    \label{fig:usd_envs}
\end{figure*}

\subsubsection{Failure diagnosis (lines~4--5)}
A scalar score cannot specify terrain, so this step exists to recover the geometry behind the number.
A single rollout would not do it either, since the location at which a policy fails varies with initialization, so each checkpoint is evaluated with $R=512$ parallel validation trials and we take the failure at the \emph{modal} traversal progress rather than the mean or the worst case, which identifies where the policy characteristically stops rather than an outlier.
The recorded context is chosen to be sufficient for authoring terrain rather than for reporting performance: the traversal distance and failure location, the termination cause, the robot base orientation, an $8\,\mathrm{m}\times8\,\mathrm{m}$ local heightmap, the terrain centerline profile, and simulator views around the failure, as shown in \Cref{fig:framework}.
Stored with the checkpoint, this describes both the policy weakness and the terrain interaction that exposed it, and it is what makes the next step possible.

\begin{algorithm}[!t]
\caption{Truncated Search-Rollout Value Estimation}
\label{alg:rollout}

\small
\begin{algorithmic}[1]

\REQUIRE Expanded policy $\pi_j$, validation terrains $\mathcal{V}$,
depth $H$, number of search rollouts $K$

\FOR{$k=1,\ldots,K$}

    \STATE $\tilde{\pi}^{(k)}_0 \leftarrow \pi_j$

    \FOR{$h=0,\ldots,H-1$}

        \STATE Evaluate $\tilde{\pi}^{(k)}_h$ on $\mathcal{V}$ to obtain
        $\mathcal{F}^{(k)}_h$

        \STATE Generate curriculum using
        $\mathcal{F}^{(k)}_h$ and rollout history

        \STATE Train to obtain $\tilde{\pi}^{(k)}_{h+1}$

    \ENDFOR

    \STATE Evaluate $\tilde{\pi}^{(k)}_H$ and record return $G^{(k)}$

\ENDFOR

\STATE Compute $V(\pi_j)$ using \Cref{eq:search_rollout_value}
\STATE Discard temporary checkpoints

\RETURN $V(\pi_j)$
\vspace{1mm}
\end{algorithmic}
\end{algorithm}

\subsubsection{Expansion (lines~6--9)}
Training only on terrain that targets the newest failure would trade an old competence for a new one, so each branch keeps a rehearsal pool of terrain it has already seen.
Only terrain that did no harm is worth keeping, so terrain from a completed expansion is retained according to the parent--child change in search value:
\begin{equation}
    \Delta Q = Q_{\mathrm{child}} - Q_{\mathrm{parent}},
    \qquad
    \begin{cases}
        \text{retain}, & \Delta Q \geq -\tau,\\
        \text{discard}, & \Delta Q < -\tau,
    \end{cases}
\end{equation}
where $\tau$ is the maximum tolerated degradation.
Each run then splits its terrain slots $50\%$ to the new curriculum and $50\%$ to rehearsal, and when the pool exceeds the slots available, recently added terrain is prioritized while the rest is retained for later reuse.
The LLM receives the failure context of $\pi_i$ together with the curriculum history along the branch leading to it, so that it does not reproduce terrain the branch has already exhausted.
It writes three components, because addressing a failure and generalizing beyond it are different requirements: a \emph{targeted} terrain that reproduces the observed failure, a \emph{generalized} terrain that varies the local geometry while preserving the underlying locomotion challenge, and a \emph{combined} terrain that incorporates both.
Together these form $\mathcal{C}_{i}$, which specifies the expansion from $s_i$ and, with its training run, defines the edge $e_{ij}$ to $s_j$.

The selected checkpoint $\pi_i$ is trained on the $50$--$50$ mixture of $\mathcal{C}_{i}$ and its rehearsal terrain, and the terrain-difficulty scheduler of the underlying parkour learner is left active throughout.
Leaving it alone is deliberate: \method changes what the learner trains on, not how it trains, so any gain is attributable to the curriculum rather than to a modified learner.
The resulting checkpoint $\pi_j$ becomes the new child node $s_j$ of $s_i$.

\subsubsection{Simulation and backpropagation (lines~10--11)}
A curriculum's worth is not visible in the checkpoint it produces immediately, because terrain that unlocks further progress may score no better than terrain that merely polishes what the policy could already do.
We therefore value a child by what can be built from it, using $K=3$ parallel \emph{search rollouts} of depth $H=2$:
\begin{equation}
    \tilde{s}_{j,0}^{(k)} = s_j
    \rightarrow
    \tilde{s}_{j,1}^{(k)}
    \rightarrow \cdots \rightarrow
    \tilde{s}_{j,H}^{(k)},
    \qquad k=1,\ldots,K ,
\end{equation}
where each transition applies the same failure-conditioned expansion the search itself uses, and the temporary checkpoints $\tilde{s}_{j,h}^{(k)}$ are discarded rather than inserted, so that value estimation does not inflate the tree.
Each temporary checkpoint is evaluated on every validation terrain with the same $R=512$ trials, and progress is again taken at the mode:
\begin{equation}
    \bar{p}_{v}^{(k)}
    =
    \operatorname{Mode}
    \left\{
        \frac{d_{v,r}^{(k)}}{L_v}
    \right\}_{r=1}^{R},
\end{equation}
with $d_{v,r}^{(k)}$ the distance traversed in trial $r$ of rollout $k$ and $L_v$ the traversable length of terrain $v$.
Averaging over the validation set gives the terminal return of a rollout:
\begin{equation}
    G^{(k)}
    =
    \frac{1}{|\mathcal{V}|}
    \sum_{v\in\mathcal{V}}
    \bar{p}_{v}^{(k)},
\end{equation}
and averaging over rollouts gives the expanded node value:
\begin{equation}
    V(s_j)
    =
    \frac{1}{K}
    \sum_{k=1}^{K}
    G^{(k)} .
    \label{eq:search_rollout_value}
\end{equation}
We propagate it with maximum rather than mean backup,
\begin{equation}
    Q(s_p,s_c)
    \leftarrow
    \max\!\left(
        Q(s_p,s_c),
        V(s_j)
    \right),
\end{equation}
for every parent--child edge $(s_p,s_c)$ on the selected path, because the search is looking for the best policy a branch can reach and a mean would penalize a branch for the weak children it also produced.
The depth $H=2$ bounds only this temporary estimate; the persistent search is anytime and runs under a wall-clock budget of $T_{\max}=24$ hours.

\begin{table}[t]
\centering
\scriptsize
\setlength{\tabcolsep}{3.5pt}
\renewcommand{\arraystretch}{1.08}

\caption{
Geometry of the \emph{relief-stratified routes} in the \textbf{Debris site benchmark}.
Values show the range across the two courses in each difficulty level.
Relief denotes local terrain height variation within a $0.4\,\mathrm{m}$
robot footprint; P95 captures the upper tail of terrain difficulty.
}
\label{tab:usd_course_stats}

\begin{tabular}{lcccc}
\toprule
\textbf{Difficulty}
& \textbf{Route (m)}
& \textbf{Mean Relief (m)}
& \textbf{P95 Relief (m)}
& \textbf{Max Step (m)}
\\
\midrule

Easy
& 11.4--20.2
& 0.128--0.144
& 0.275--0.295
& 0.26--0.32
\\

Mid
& 10.4--15.4
& 0.204--0.210
& 0.343--0.389
& 0.32--0.43
\\

Hard
& 9.6--19.2
& 0.290--0.303
& 0.459--0.503
& 0.58--0.76
\\

V-Hard
& 13.7--17.8
& 0.318--0.325
& 0.589--0.629
& 0.51--0.87
\\

\bottomrule
\end{tabular}
\end{table}

\subsubsection{Parallel search}
Since each expansion requires a full training run, we execute multiple tree
expansions asynchronously following parallel MCTS principles~\cite{parallelmcts}.
Rather than using virtual loss, which would perturb the branch value
$Q(s_i,s_j)$, we maintain an in-flight count $N_{\mathrm{inf}}(s)$ for active
expansions and incorporate it only into the visit counts,
\begin{equation}
    S_{\mathrm{par}}(s_i,s_j)\!
    =\!
    Q(s_i,s_j)
    +
    c_{\mathrm{puct}}
    \frac{
        \sqrt{N(s_i)+N_{\mathrm{inf}}(s_i)}
    }{
        1+N(s_j)+N_{\mathrm{inf}}(s_j)
    }.
\end{equation}

%% file: sections/experiments.tex
\section{EXPERIMENTS}
\label{sec:experiments}

\textbf{Baselines.} We compare against Extreme Parkour (EP)~\cite{extremeparkour},
Eurekaverse~\cite{eurekaverse}, and a modified Eurekaverse baseline that we
denote Eurekaverse + Validation Reference. EP uses the original
hand-designed parkour terrain curriculum. For Eurekaverse, we follow the
original evolutionary terrain-generation procedure, using Claude Opus 4.8 in
place of the GPT-4o used in the original work. To construct Modified Eurekaverse, we modify
Eurekaverse so that its terrain generator additionally receives the same
validation-terrain references and prompting used to specify the validation
domain in \method. All other aspects of the Eurekaverse generation procedure are
kept unchanged. Modified Eurekaverse therefore controls for the benefit of exposing the LLM to validation-domain information without failure-conditioned feedback or curriculum search.

\textbf{Real-World Reconstructed Environments.}
We evaluate on two real-world reconstructed environments: \textbf{Debris site and Exposed Beam site}, from complex industrial and
disaster-site geometry, shown in~\Cref{fig:usd_envs}. These worlds, in particular, contain challenges such as
irregular elevation changes, discontinuities, and
compound geometric structures. These sites are held out from
\method curriculum generation and policy training.

\begin{table}[t]
\centering
\vspace{2.5mm}
\scriptsize
\setlength{\tabcolsep}{3.0pt}
\renewcommand{\arraystretch}{0.95}
\setlength{\aboverulesep}{1pt}
\setlength{\belowrulesep}{1pt}
\setlength{\abovecaptionskip}{3pt}
\setlength{\belowcaptionskip}{2pt}

\caption{
Performance on \textit{relief-stratified routes} in the \textbf{Debris site},
with two routes per difficulty tier.
Each route is evaluated over two independent seeds with 256 trials per seed.
WP denotes median waypoint completion normalized by the number of route waypoints.
Higher is better ($\uparrow$).
}
\label{tab:debris_difficulty}

\begin{tabular}{@{}llccc@{}}
\toprule
\textbf{Difficulty}
& \textbf{Method}
& \textbf{Success (\%) $\uparrow$}
& \textbf{Progress IQM $\uparrow$}
& \textbf{WP (\%) $\uparrow$}
\\
\midrule

Easy
& Extreme Parkour
& 29.8
& 0.77
& 74.1
\\
& Eurekaverse
& 29.4
& 0.45
& 26.8
\\
& Modified Eurekaverse
& 31.5
& 0.61
& 55.4
\\
& \textbf{\method (Ours)}
& \textbf{84.0}
& \textbf{1.00}
& \textbf{100.0}
\\

\midrule

Mid
& Extreme Parkour
& 23.9
& 0.63
& 53.2
\\
& Eurekaverse
& 28.4
& 0.70
& 67.5
\\
& Modified Eurekaverse
& 28.4
& 0.70
& 60.3
\\
& \textbf{\method (Ours)}
& \textbf{85.8}
& \textbf{1.00}
& \textbf{100.0}
\\

\midrule

Hard
& Extreme Parkour
& 11.8
& 0.41
& 36.6
\\
& Eurekaverse
& 0.8
& 0.34
& 31.4
\\
& Modified Eurekaverse
& 14.9
& 0.42
& 39.5
\\
& \textbf{\method (Ours)}
& \textbf{44.7}
& \textbf{0.82}
& \textbf{79.4}
\\

\midrule

V-Hard
& Extreme Parkour
& 0.6
& 0.34
& 26.0
\\
& Eurekaverse
& 0.2
& 0.44
& 36.5
\\
& Modified Eurekaverse
& 0.8
& 0.39
& 25.0
\\
& \textbf{\method (Ours)}
& \textbf{11.4}
& \textbf{0.71}
& \textbf{64.6}
\\

\bottomrule
\end{tabular}

\end{table}

\textbf{Evaluation Routes.}
Each evaluation route begins from a prescribed robot spawn location and follows
a fixed sequence of waypoints through the reconstructed site. All evaluation routes are fixed before policy evaluation. We consider two construction procedures:
\emph{relief-stratified routes}, whose paths are generated from terrain geometry
using a relief-based A* procedure, and \emph{directly specified routes}, whose
paths are defined without the relief-based generator.

\textbf{Debris site.}
For the Debris site, we use terrain geometry to automatically construct routes
at different difficulty levels. We first compute a terrain-relief map, where
\emph{relief} measures the local vertical variation within a
$0.4\,\mathrm{m}$ footprint, approximately matching the spatial scale of the
robot. Low relief corresponds to locally smooth terrain, while larger values
capture stronger elevation changes, edges, steps, and other non-planar
geometry.

We define target relief ranges for each difficulty level and use A* to generate
routes from the prescribed spawn location through regions matching the desired
range. Each resulting path is converted into evaluation waypoints, and routes
whose measured relief falls outside the target range are rejected. This produces
eight evaluation routes, with two routes each at \emph{Easy}, \emph{Mid},
\emph{Hard}, and \emph{V-Hard} difficulty. The geometric characteristics of the evaluation routes are summarized in
Table~\ref{tab:usd_course_stats}.

\textbf{Exposed Beam site.}
The Exposed Beam site contains large gaps, narrow structural members, and
disconnected regions for which local relief alone is not a reliable measure of
route difficulty. We therefore use fixed evaluation routes through the
traversable structure and group them into \emph{Mid} and \emph{Hard} settings.

\begin{figure*}[ht]
    \centering
    \vspace{2.5mm}
\includegraphics[width=\textwidth, height=0.3\textheight, keepaspectratio]{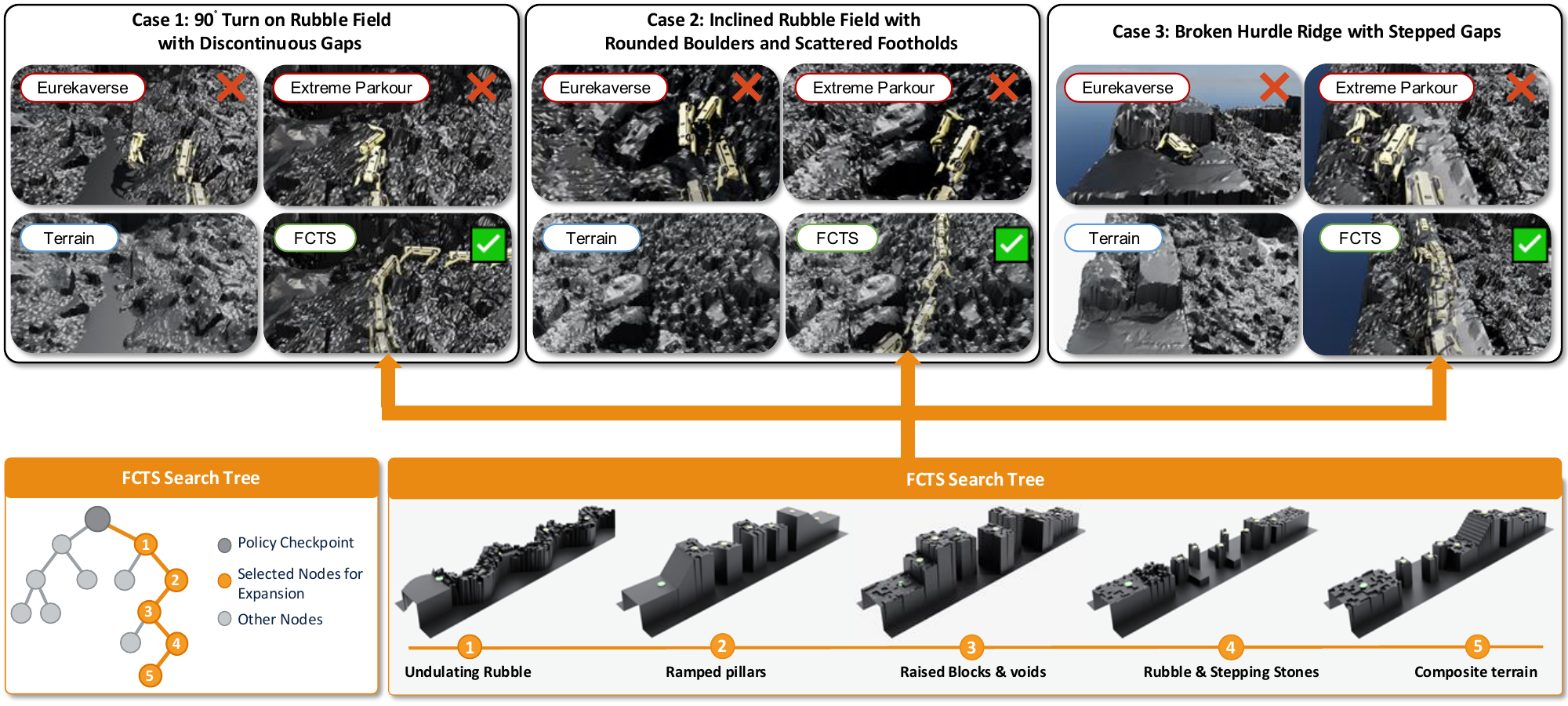}    \caption{
    Failure cases and the corresponding winning FCTS curriculum branch. Top: three
terrain regions where Extreme Parkour and Eurekaverse fail while FCTS completes
the traversal. Bottom: the highest-valued branch found by FCTS and the terrain
families generated along it. The branch progressively introduces irregular
rubble, inclined support, repeated height-and-gap transitions, and their
composition.
    }
    \label{fig:failure_analysis}
\end{figure*}

\textbf{Evaluation Protocol.}
Each route is evaluated over two seeds with 256 parallel robots per seed, for
512 trials in total. All methods share the same routes, spawn locations,
$0.6\,\mathrm{m/s}$ command, $100\,\mathrm{s}$ episode limit, and
domain-randomization settings; initial robot state, including base orientation,
varies across trials.

We report three metrics: \emph{Success Rate}, the fraction of trials reaching
the final waypoint; \emph{Progress IQM}, the interquartile mean of normalized
route completion; and \emph{Waypoint Progress}, the median waypoint reached
normalized by the total number of waypoints.

\begin{table}[t]
\centering
\vspace{2.5mm}
\scriptsize
\setlength{\tabcolsep}{3.2pt}
\renewcommand{\arraystretch}{0.98}
\setlength{\aboverulesep}{1pt}
\setlength{\belowrulesep}{1pt}
\setlength{\abovecaptionskip}{3pt}
\setlength{\belowcaptionskip}{2pt}

\caption{
Performance on \textit{directly specified routes} in the \textbf{Debris site and
Exposed Beam Site}.
Each route is evaluated over two independent seeds with 256 trials per seed.
WP denotes median waypoint completion normalized by the number of route waypoints.
Higher is better ($\uparrow$).
}
\label{tab:site_routes}

\begin{tabular}{@{}llccc@{}}
\toprule
\textbf{Difficulty}
& \textbf{Method}
& \textbf{Success (\%) $\uparrow$}
& \textbf{Progress IQM $\uparrow$}
& \textbf{WP (\%) $\uparrow$}
\\
\midrule

\multicolumn{5}{@{}l}{\textbf{Debris site}} \\
\cmidrule(lr){1-5}

Easy
& Extreme Parkour
& 0.8
& 0.27
& 30.0
\\
& Eurekaverse
& 2.3
& 0.31
& 40.0
\\
& Modified Eurekaverse
& 1.2
& 0.21
& 20.0
\\
& \textbf{\method (Ours)}
& \textbf{36.7}
& \textbf{0.59}
& \textbf{50.0}
\\

\cmidrule(lr){1-5}

Mid
& Extreme Parkour
& 5.1
& 0.16
& 0.0
\\
& Eurekaverse
& 0.0
& 0.22
& 0.0
\\
& Modified Eurekaverse
& 2.7
& \textbf{0.36}
& 0.0
\\
& \textbf{\method (Ours)}
& \textbf{23.8}
& 0.33
& \textbf{33.3}
\\

\cmidrule(lr){1-5}

Hard
& Extreme Parkour
& 0.4
& 0.19
& 0.0
\\
& Eurekaverse
& 0.0
& 0.19
& 0.0
\\
& Modified Eurekaverse
& 3.5
& 0.28
& 20.0
\\
& \textbf{\method (Ours)}
& \textbf{41.4}
& \textbf{0.71}
& \textbf{60.0}
\\

\midrule

\multicolumn{5}{@{}l}{\textbf{Exposed Beam Site}} \\
\cmidrule(lr){1-5}

Mid
& Extreme Parkour
& 0.0
& 0.07
& 0.0
\\
& \textbf{Eurekaverse}
& \textbf{25.0}
& \textbf{0.54}
& \textbf{45.5}
\\
& Modified Eurekaverse
& 3.5
& \textbf{0.54}
& \textbf{45.5}
\\
& \method (Ours)
& 10.2
& 0.44
& \textbf{45.5}
\\

\cmidrule(lr){1-5}

Hard
& Extreme Parkour
& 10.9
& 0.37
& 28.6
\\
& Eurekaverse
& 19.1
& 0.37
& 28.6
\\
& Modified Eurekaverse
& 23.8
& \textbf{0.48}
& \textbf{42.9}
\\
& \textbf{\method (Ours)}
& \textbf{27.7}
& 0.42
& 28.6
\\

\bottomrule
\end{tabular}

\end{table}

\textbf{Portfolio-Based Checkpoint Selection.}
A central objective of this work is reliable policy deployment on real-world
industrial and disaster-site terrain. \method therefore retains the checkpoints
discovered across different curriculum trajectories as a policy portfolio.
Given a target site's reconstructed USD terrain, we evaluate this portfolio and
select the checkpoint with the highest traversal performance on that terrain.

\subsection{Performance on Real-World Reconstructed Environments}
\label{sec:generalization}

\Cref{tab:debris_difficulty} and \Cref{tab:site_routes} summarize our results. \method maintains a clear advantage across increasing terrain difficulty. On the
Easy tier, it reaches $84.0\%$ success, compared with $31.5\%$ for the strongest
baseline, and achieves $100\%$ median waypoint completion. The gap remains large
on harder terrain: \method reaches $44.7\%$ success with a Progress IQM of $0.82$
on Hard, compared with $14.9\%$ success and $0.42$ IQM for the strongest
baseline. On V-Hard, \method reaches $11.4\%$ success while all baselines remain
below $1\%$, with a Progress IQM of $0.71$ compared with at most $0.44$.

The manually specified routes show a similar advantage on Debris site, where \method
achieves the highest success rate on all three routes. Exposed Beam site is more mixed:
Eurekaverse performs best on Route 1 with $25.0\%$ success compared with
$10.2\%$ for \method, while \method performs best on Route 2 with $27.7\%$ success
compared with $23.8\%$ for the next-best method.

\subsection{Why Does FCTS Succeed Where the Baselines Fail?}
\label{sec:why_fcts_succeeds}

FCTS achieves the \textbf{largest gains on the most difficult evaluation routes}, where
the success-rate gap over the baselines is much larger than the gap in traversal
progress. This suggests that the difference is concentrated at a small number of
critical terrain regions rather than spread uniformly across the course. We
therefore inspect representative failure locations where Extreme Parkour and
Eurekaverse terminate while FCTS continues, and trace these successes back to
the curriculum trajectory discovered by FCTS.

\Cref{fig:failure_analysis} shows three such cases together with
the strongest branch of the final search tree. We define the \textbf{\emph{winning
branch}} by starting at the root and recursively following the child with the
highest backed-up $Q$ value. Since FCTS uses max backup, this path leads to the
highest-valued descendant found during search. The branch contains, in order,
\emph{Undulating Rubble}, \emph{Ramped Pillars}, \emph{Raised Blocks \& Voids},
\emph{Heterogeneous Rubble Field}, and a final \emph{Composite Terrain}. None of
the evaluation scenes are used during curriculum generation.

We make three observations. First, \textbf{FCTS succeeds at the $90^\circ$ turn where
both baselines fail}. At this location, the turn coincides with uneven rubble and
gaps in the traversable surface. The FCTS branch starts with strongly varying
surface geometry and later introduces larger changes in support height and
spacing before reaching the composite terrain. In the evaluation trajectory,
FCTS completes the turn while both baselines terminate in this region.

Second, the largest behavioral difference is visible on the inclined section
with rounded boulders and scattered footholds. Both baseline policies develop
large lateral roll and then trip or fall. \textbf{FCTS remains upright and crosses the
same incline}. The branch introduces \emph{Ramped Pillars} early in the search, which
adds sustained slope and non-uniform support to the training curriculum. Later
nodes further vary support height and placement. The final FCTS policy therefore
reaches this case after several expansions that progressively increase the
difficulty of maintaining stability on irregular inclined support.

Third, \textbf{FCTS crosses the broken hurdle sequence while both baselines fail} at the
repeated gap transitions. The relevant curriculum change occurs at
\emph{Raised Blocks \& Voids}, which introduces consecutive elevated surfaces
separated by gaps. Later nodes retain this terrain and add further variation
before the final composite curriculum. FCTS consequently completes several
successive height-and-gap transitions in the evaluation course, whereas the
baseline trajectories terminate within the sequence.

\subsection{Parallel Search Efficiency}
\label{sec:parallel_efficiency}

Each \method expansion requires a complete policy-training run, which makes sequential
tree expansion expensive in wall-clock time. Table~\ref{tab:parallel_efficiency}
compares sequential and asynchronous execution under the same $24$-hour search
budget. Parallel search increases the expansion rate from $0.21$ to
$1.13$ completed expansions per hour, and increases the number of completed
expansions from approximately $5.1$ to $25$ and shows a
$5.3\times$ improvement in search throughput.

\begin{figure}[!t]
    \centering
    \vspace{2.5mm}
    \includegraphics[
        width=0.9\columnwidth,
        height=0.3\textheight,
        keepaspectratio
    ]{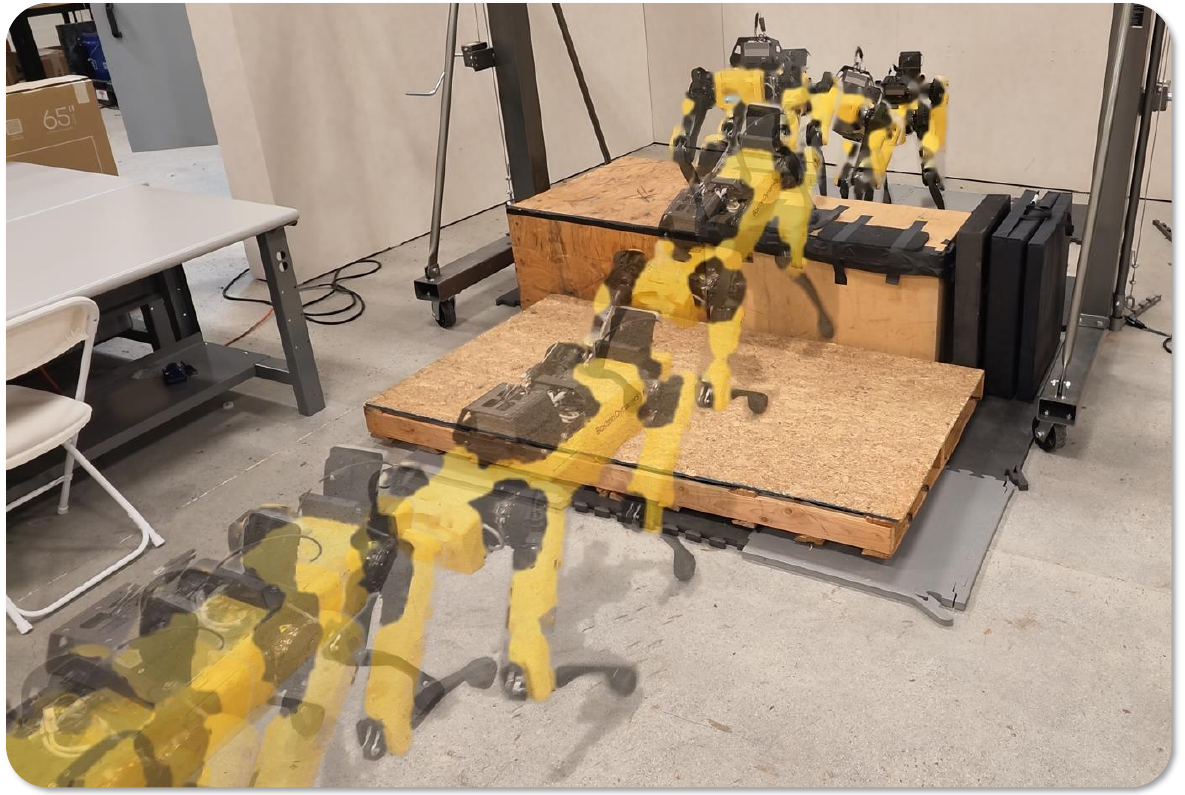}
    \caption{Real-world deployment of FCTS on a Boston Dynamics Spot traversing
    a compound parkour course consisting of a pallet, raised block, and stair descent.}
    \label{fig:real_robot}
\end{figure}

\begin{table}[t]
\centering
\caption{Computational efficiency of parallel curriculum search.}
\label{tab:parallel_efficiency}
\begin{tabular}{lcccc}
\toprule
\textbf{Method}
& \textbf{GPU Util.}
& \textbf{Exp./hr}
& \textbf{24h Exp.}
& \textbf{Speedup}
\\
\midrule
Sequential MCTS
& $28\%$ & $0.21$ & $5.1$ & $1.0\times$ \\
Parallel MCTS
& $20\%$ & $\bm{1.13}$ & $\bm{25}$ & $\bm{5.3\times}$ \\
\bottomrule
\end{tabular}
\end{table}

\subsection{Real-World Transfer}
\label{sec:hardware}

We evaluate sim-to-real transfer on a Boston Dynamics Spot using a compound
physical parkour course comprising a pallet, an elevated block, and a stair
descent. We deploy the FCTS policy at 50\,Hz on an onboard NVIDIA Jetson AGX
Orin, using depth observations from an Intel RealSense D435i together with robot
proprioception. Figure~\ref{fig:real_robot} shows the resulting hardware
traversal, which shows that the FCTS policy can be
successfully transferred to the physical robot.

%% file: sections/conclusions.tex
\section{CONCLUSIONS}

We presented \method, a failure-conditioned curriculum framework that uses
localized validation failures to generate new training terrains and MCTS to
explore multiple curriculum trajectories. This allows the search to
expand different policy checkpoints instead of committing to a single sequence
of terrain updates. On held-out evaluation routes in reconstructed real-world
environments, \method improves both traversal success and progress over baselines. We also demonstrate transfer to a Boston Dynamics Spot on a
compound physical parkour course. Future work will study deployment on real industrial sites.